\documentclass[letterpaper, 10 pt, conference]{ieeeconf}  

\IEEEoverridecommandlockouts                              

\usepackage{graphicx} 
\usepackage{mathtools}
\usepackage{amsmath} 
\usepackage{amssymb}  
\usepackage{mathrsfs}  
\usepackage{cuted}
\usepackage{color}
\usepackage{subcaption}
\usepackage[binary-units=true]{siunitx}
\usepackage{lipsum}  
\usepackage{romannum}
\usepackage{hyperref}

\usepackage[bordercolor=white,backgroundcolor=gray!30,linecolor=black,colorinlistoftodos]{todonotes}

\newcommand{\lambdacstr}{\bar{\lambda}}
\newcommand{\windratio}{\beta}
\newcommand{\feas}{\operatorname{feas}}
\newcommand{\vgco}{v_{G,\text{co}}}
\newcommand{\vgmin}{v_{G,\text{min}}}
\newcommand{\vgz}{v_{G_0}}
\newcommand{\Deltavamax}{\Delta v_{A,\text{max}}}
\newcommand{\Deltavaw}{\Delta v_A^w}
\newcommand{\Deltavae}{\Delta v_A^e}
\newcommand{\Deltavaemax}{\Delta v_{A,\text{max}}^e}
\newcommand{\sat}{\operatorname{sat}}
\newcommand{\eb}{e_b}
\newcommand{\enorm}{\bar{e}}
\newcommand{\Deltaebuf}{\enorm_\text{buf}}
\newcommand{\Deltawbuf}{\Delta w_\text{buf}}
\newcommand{\varef}{v_{A,\text{ref}}}
\newcommand{\vanom}{v_{A,\text{nom}}}
\newcommand{\vamax}{v_{A,\text{max}}}
\newcommand{\curv}{\kappa_P}
\newcommand{\course}{\chi}
\newcommand{\bearing}{\course_\text{ref}}
\newcommand{\heading}{\xi}
\newcommand{\headingref}{\heading_\text{ref}}
\newcommand{\lookahead}{\hat{\mathbf{l}}}
\newcommand{\lookaheadA}{\lookahead_A}
\newcommand{\lookaheadAz}{\lookaheadA^{\curv=0}}
\newcommand{\lookaheadAfeas}{\lookahead_{A,\text{feas}}}
\newcommand{\lookaheadAinfeas}{\lookahead_{A,\text{infeas}}}
\newcommand{\errorangleG}{\eta}
\newcommand{\errorangleA}{\eta_A}

\title{\LARGE \bf
On Flying Backwards: Preventing Run-away of Small, Low-speed, Fixed-wing UAVs in Strong Winds
}

\author{Thomas Stastny and Roland Siegwart
\thanks{All authors are with the Autonomous Systems Lab at the Swiss Federal Institute of Technology (ETH Z\"{u}rich), Z\"{u}rich, Switzerland.
        {\tt\small thomas.stastny@mavt.ethz.ch}}
}

\begin{document}

\begin{minipage}{\textwidth}
\copyright 2019 IEEE. Personal use of this material is permitted. Permission from IEEE must be obtained for all
other uses, in any current or future media, including reprinting/republishing this material for advertising
or promotional purposes, creating new collective works, for resale or redistribution to servers or lists, or
reuse of any copyrighted component of this work in other works.\\

Please cite this paper as:\\%
\begin{verbatim}
@inproceedings{stastny2019_onflyingbackwards,
  title     = "On Flying Backwards: Preventing Run-away of Small, Low-speed,
               Fixed-wing UAVs in Strong Winds",
  author    = "Stastny, Thomas and Siegwart, Roland",
  booktitle = "2019 {IEEE/RSJ} International Conference on Intelligent Robots
               and Systems ({IROS})",
  year      = 2019;
}
\end{verbatim}

\end{minipage}

\maketitle
\thispagestyle{empty}
\pagestyle{empty}

\begin{abstract}
Small, low-speed fixed-wing Unmanned Aerial Vehicles (UAVs) operating autonomously, beyond-visual-line-of-sight (BVLOS) will inevitably encounter winds rising to levels near or exceeding the vehicles' nominal airspeed.
In this paper, we develop a nonlinear lateral-directional path following guidance law with explicit consideration of online wind estimates.
Energy efficient airspeed reference compensation logic is developed for excess wind scenarios (i.e. when the wind speed rises above the airspeed), enabling either mitigation, prevention, or over-powering of excess wind induced run-away from a given path.
The developed guidance law is demonstrated on a representative small, low-speed test UAV in two flight experiments conducted in mountainous regions of Switzerland with strong, turbulent wind conditions, gusts reaching up to 13 meters per second.
We demonstrate track-keeping errors of less than 1 meter consistently maintained during a representative duration of gusting, excess winds and a mean ground speed undershoot of 0.5 meters per second from the commanded minimum forward ground speed demonstrated in over 5 minutes of the showcased flight results.
\end{abstract}

\section{INTRODUCTION}
In recent years, small, easily manageable, operated, and maintained fixed-wing Unmanned Aerial Vehicles (UAVs) are increasingly being applied to remote sensing ventures requiring long-range and/or long-endurance flight.
For example, ETH Z\"{u}rich's Autonomous Systems Lab (ASL) has developed Low-Altitude, Long-Endurance (LALE) solar-powered platforms capable of multi-day, payload-equipped flight~\cite{oettershagen_jfr2018}, and further demonstrated the utility of such small platforms in beyond-visual-line-of-sight (BVLOS) science missions such as Arctic glacier monitoring (see project Sun2Ice\footnote{\url{http://www.sun2ice.ethz.ch}}).
The ability to reach far-away locations where humans either cannot or do not want to go is a great advantage, however it also comes with risks that operators, placated by seemingly fully automated aircraft, may not anticipate.
Endemic to small, low-speed fixed-wing platforms is a susceptibility to high winds.
UAVs operating autonomously BVLOS, e.g. in mountainous areas or along coastlines, will  doubtless encounter winds rising to levels near or exceeding the vehicles' nominal airspeed.
Without a control law cognizant of the local wind field, or moreover, without logic to handle such cases, the aircraft risks mission delays (unable to make meaningful progress towards subsequent waypoints) or possibly loss of airframe (if operating near and blown into large structures, e.g. cliff walls or fjords).
As researchers and practitioners execute more flight hours and experience more weather conditions, wind hazards have started to be acknowledged within literature.
Recent work has considered wind in various iterations of emergency landing planners using airmass relative Dubins aircraft curves (or trochoids in the inertial frame)~\cite{warren2015_emergencylanding, Klein2017_emergencylanding_dubinswind}, however, these planners still require that the aircraft is able to move forward with respect to the ground and rely on some form of guidance law to follow the planned paths.
Authors of~\cite{Marina2017_vectorfieldpaparazi} propose a vectorfield -based guidance law which considers the current wind estimate and may follow any smooth path curvature, however the algorithm uses level sets to describe the notion of distance from the trajectory which makes tuning specific to the curve for which the ``distance" function is defined, as opposed to the more common direct relation to track error. 
A three-dimensional guidance approach with explicit consideration of wind is developed in~\cite{beard2014pathfollowwind}, where the law further accounts for roll and flight path angle constraints using the theory of nested saturations.
\begin{figure}[t!]
\centering
\includegraphics[width=\linewidth]{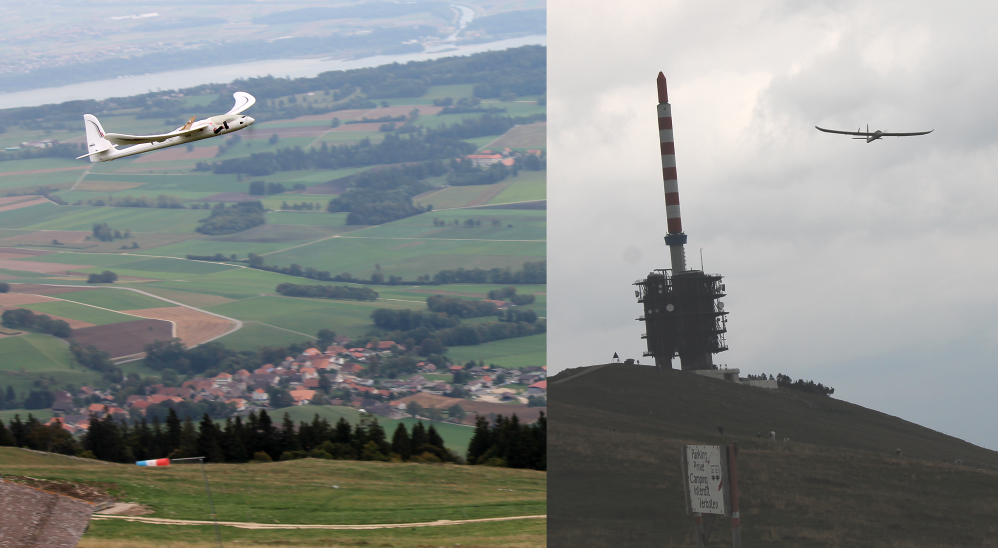}
\caption{Easyglider test platform flying in strong winds over the Jura, Switzerland.}
\label{fig:teaser}
\end{figure}
More generic guidance logic for either waypoint tracking or path following typically takes inertial ground speed measurements as inputs, which contains the effect of wind.
Perhaps the most widely used path following guidance running on small fixed-wing UAVs today, the nonlinear path following guidance developed in~\cite{Park2007_stability} (commonly known as $L_1$ guidance), uses exactly this approach.
This ``look-ahead" method is simple (implemented easily on a microcontroller), intuitive to tune (following further extentions in~\cite{Curry2013_L2plus}), and quite effective in practice; however, it has the detriments that 1) convergence to the path is only guaranteed for lines and circles, and 2) as shown in~\cite{Furieri2017_windyNPFG}, the logic breaks down once winds approach and/or exceed the vehicle's airspeed.
The law was abstracted in~\cite{Cho2016_NPFG} to a more general form capable of following any smooth, continuous 3D path, though without consideration of wind.
In our previous work, we took the method from~\cite{Cho2016_NPFG} in a different direction and reformulated the 2D case for consideration of \emph{excess wind} cases (i.e. wind speed $>$ airspeed)~\cite{Furieri2017_windyNPFG}.
To date,~\cite{Furieri2017_windyNPFG} is still the only guidance method in literature considering the particular problem of \emph{excess wind}.
While this logic will provide run-away \emph{mitigation} (i.e. minimizing the rate at which the vehicle is blown away from the path), in the case the aircraft may have remaining energy available, the airspeed reference could be increased above the nominal value to further reduce, or even \emph{prevent} run-away.
In this work, we propose utilizing the airspeed reference as control towards the development of an efficient airspeed reference compensation logic, running in parallel with an improved, wind-robust directional guidance.
The resultant logic either regulates wind excess, stays on track, or maintains a minimum ground speed, depending on the operator's chosen mode or the aircraft's speed limits.
We provide significant enhancements to the baseline algorithm with a heavy emphasis on practical implications of fielding the controller including, but not limited to, an improved notion of \emph{bearing feasibility}, numerical stability considerations, reference command continuity, and condition independent tuning strategies.
Finally, we demonstrate the effectiveness of the control strategy on a small fixed-wing test platform in thorough flight experiments in mountainous terrain with strong, turbulent winds.
The resulting guidance is, to the authors' best knowledge, the first example in literature of an algorithm considering both excess wind conditions on small fixed-wing UAVs as well as providing the means to fully prevent vehicle run-away and maintain track keeping.
The remainder of the paper is structured as follows: Section~\ref{sec:bearing_feasibility} presents the objective formulation and the concept of \emph{bearing feasibility}, Section~\ref{sec:dir_guidance} outlines our previously developed directional guidance strategy~\cite{Furieri2017_windyNPFG} for both lower and \emph{excess} wind conditions, detailing specific enhancements and modifications in the present work, Section~\ref{sec:airspeed_reference_compensation} develops a new airspeed reference compensation logic, and Section~\ref{sec:flight_experiments} concludes with experimental results.
%

\section{BEARING FEASIBILITY}
\label{sec:bearing_feasibility}
\begin{figure}
\centering
\includegraphics[width=0.7\linewidth,trim={0 0.45cm 0 0},clip]{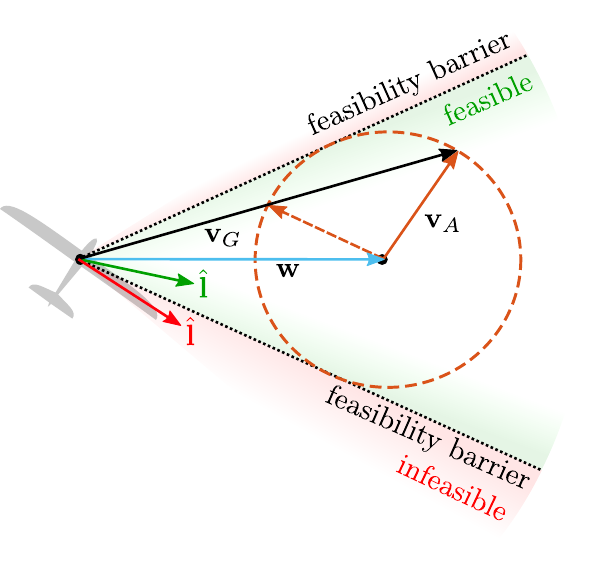}
\caption{Feasibility ``cone" (wind speed greater than airspeed). Note in the \emph{excess wind} condition, two heading solutions exist for a given course.}
\label{fig:feas_cone}
\end{figure}

\begin{figure}
\centering
\includegraphics[width=\linewidth]{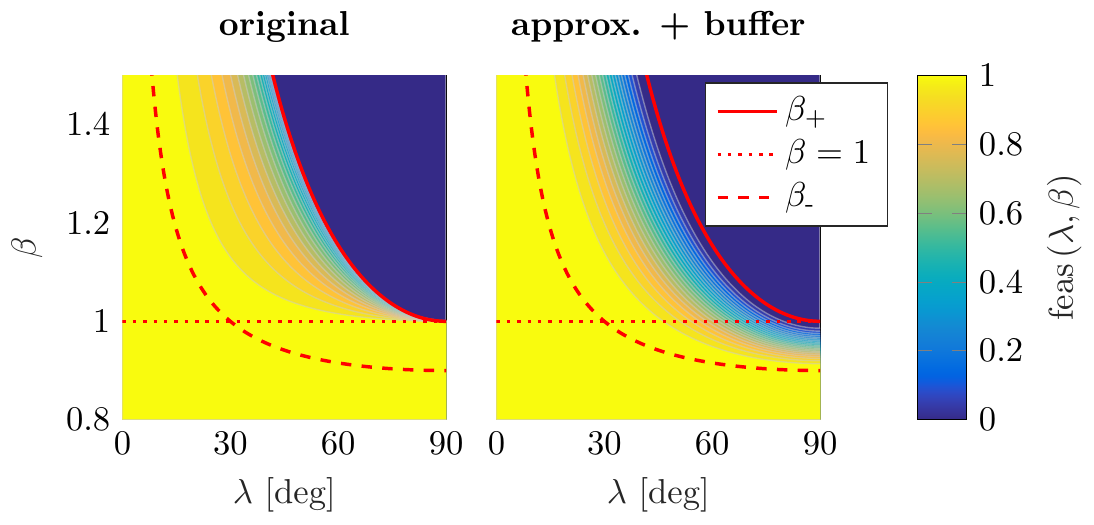}
\caption{Feasibility function: original formulation from~\cite{Furieri2017_windyNPFG} (left), new approximation with extended buffer zone (right).}
\label{fig:feas_func}
\end{figure}

We consider a fixed-wing aircraft flying horizontally in two dimensions with ground velocity $\mathbf{v}_G = \mathbf{v}_A + \mathbf{w}$ the sum of airspeed $\mathbf{v}_A$ and wind $\mathbf{w}$ vectors.
In the case that the wind speed exceeds the UAV's airspeed, feasibility of flying a given bearing $\bearing$ depends on the wind direction.
A binary definition of the \emph{bearing feasibility} can be formulated as:
\begin{equation}\label{eq:feas_binary}
\begin{array}{ll}
\windratio\sin|\lambda|\geq 1 \cup \left(|\lambda|\geq\frac{\pi}{2} \cap \windratio>1\right) & \text{(infeasible)} \\
\text{else} & \text{(feasible)} \\
\end{array}
\end{equation}
\noindent where the wind ratio $\windratio=w/v_A$ is the fraction of wind speed $w=\|\mathbf{w}\|$ over airspeed $v_A = \|\mathbf{v}_A\|$ and $\lambda$ is the angle between the wind $\mathbf{w}$ and (unit) bearing $\lookahead=\left[\cos\bearing, \sin\bearing\right]^T$ vectors:
\begin{equation}\label{eq:lambda}
\lambda = \operatorname{atan2}\left(\mathbf{w}\times\lookahead,\mathbf{w}\cdot\lookahead\right)\in\left[-\pi,\pi\right]
\end{equation}
The relationship in~\eqref{eq:feas_binary} physically describes a ``feasibility cone", fully open when $\windratio<1$ and asymptotically decreasing to zero angular opening as $\windratio\rightarrow\infty$, see Fig.~\ref{fig:feas_cone}.
When the bearing $\lookahead$ lies within this cone the bearing is \emph{feasible} and contrarily, when outside, \emph{infeasible}.
Two separate tracking objectives can then be intuited: 1) an \emph{ideal} tracking objective, where we are able to track the prescribed bearing and 2) a \emph{safety} objective, where we instead tend towards reducing run-away by turning against the wind and simultaneously leveling the aircraft as $t\rightarrow \infty$, where $t$ is time.
When the vehicle remains on or near the feasibility boundary (common when the wind speed is approaching the airspeed and small gusts or turbulence are present), it is desirable to transition continuously between these two states to avoid oscillating discretely between reference commands (see Section~\ref{sec:dir_guidance} for reference command generation).
In~\cite{Furieri2017_windyNPFG}, the following continuous feasibility function was proposed:
\begin{equation}\label{eq:feas_old}
\feas\left(\lambda,\windratio\right) = \frac{\sqrt{1-\left(\windratio\sin\lambdacstr\right)^2}}{\cos\lambdacstr}
\end{equation}
\noindent where $\feas\left(\lambda,\windratio\right)\in\left[0,1\right]$ transitions from a value of $1$ at ``fully" feasible conditions ($\windratio<1$) to $0$ in infeasible conditions (definition in \eqref{eq:feas_binary}), see Fig.~\ref{fig:feas_func} (left).
Input $\lambdacstr=\sat\left(|\lambda|,0,\frac{\pi}{2}\right)$, where operator $\sat\left(\cdot,\operatorname{min},\operatorname{max}\right)$ saturates the input at the bounds $\operatorname{min}$ and $\operatorname{max}$.
However, some practical issues exist with the function as defined in~\eqref{eq:feas_old}; namely:
\begin{itemize}
\item The function is continuous, but not smooth at the feasibility boundary, which can lead to fast changing and undesirably jagged reference commands.
\item Numerical stability issues exist as $\lambdacstr\rightarrow\frac{\pi}{2} \cap \windratio\rightarrow 1$ due to the simultaneously decreasing magnitudes of the numerator and denominator (calculations with floating point precision on small microcontrollers then become an issue).
\item A purely binary jump from feasible to infeasible conditions exists at $|\lambda|\geq\frac{\pi}{2} \cap \windratio=1$, which leads to jumping reference commands at a critical and common position in the state space: i.e. when the wind speed is very close the airspeed, the aircraft is facing against the wind $\lambda=\pi$, and small gusts perturb the system above and below the feasibility barrier.
\end{itemize}
To address these issues, a small buffer zone below the $\windratio=1$ line is designed, considering some buffering wind ratio $\windratio_\text{buf}\in\left(0,1\right)$.
\noindent The buffer's magnitude may be set e.g. corresponding to the airspeed reference tracking dynamics.
An approximation of the feasibility function in~\eqref{eq:feas_old} can be made incorporating the buffer zone, as well as maintaining both continuity and smoothness in the transition (see Fig.~\ref{fig:feas_func} (right)):
\begin{equation}\label{eq:feas_new}
\feas\left(\lambda,\windratio\right)= \begin{cases}
0 & \windratio > \windratio_+ \\
\cos^2\left(\frac{\pi}{2}\sat\left(\frac{\windratio-\windratio_-}{\windratio_+-\windratio_-},0,1\right)\right) & \windratio > \windratio_- \\
1 & \text{else}
\end{cases}
\end{equation}
where the upper limit of the transitioning region $\windratio_+$ is approximated as a piecewise-continuous function with a linear finite cut-off to avoid singularities, the cut-off angle $\lambda_\text{co}$ chosen small such that the regular operational envelope is not affected:
\begin{equation}
\windratio_+ = \begin{cases}
\windratio_{+_\text{co}} + m_\text{co}\left(\lambda_\text{co}-\lambdacstr\right) & \lambdacstr < \lambda_\text{co} \\
1/\sin\lambdacstr & \text{else}
\end{cases}
\end{equation}
\noindent with $\windratio_{+_\text{co}} = 1/\sin\lambda_\text{co}$ and $m_\text{co} = \cos\lambda_\text{co}/\sin\lambda_\text{co}^2$. The lower limit of the transitioning region $\windratio_-$ is similarly made piecewise-continuous to correspond with $\windratio_+$:
\begin{equation}
\windratio_- = \begin{cases}
\windratio_{-_\text{co}} + m_\text{co}\left(\lambda_\text{co}-\lambdacstr\right)\windratio_\text{buf} & \lambdacstr < \lambda_\text{co} \\
\left(1/\sin\lambdacstr-2\right)\windratio_\text{buf} + 1 & \text{else} 
\end{cases}
\end{equation}
\noindent where $\windratio_{-_\text{co}} = \left(1/\sin\lambda_\text{co}-2\right)\windratio_\text{buf} + 1$.
%

\section{DIRECTIONAL GUIDANCE}
\label{sec:dir_guidance}
\begin{figure}
\centering
\includegraphics[width=0.9\linewidth]{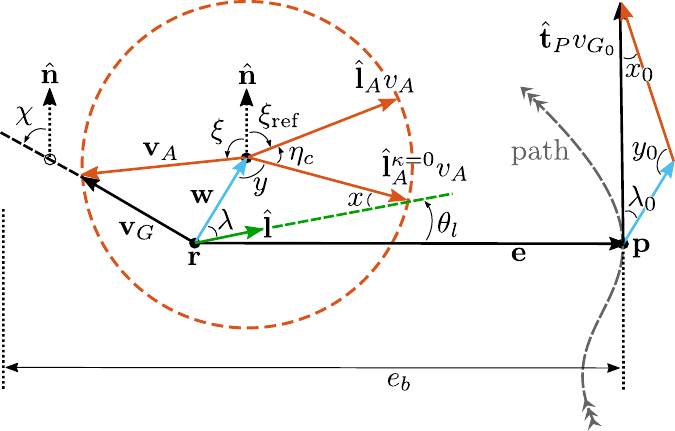}
\caption{Directional guidance -- geometry for feasible bearing.}
\label{fig:geom}
\end{figure}

For purely directional guidance, traditional look-ahead approaches (\cite{Park2007_stability, Cho2016_NPFG}) consider a constant speed unicycle model directionally driven via normal acceleration command
\begin{equation}
a_{G,\text{ref}}^N = k{v_G}^2\sin\errorangleG
\end{equation}
typically defined about the ground speed vector, where $k$ is a proportional gain and $\errorangleG=\bearing-\course \in\left[-\pi,\pi\right]$ is the angular error in course $\course$ from the bearing $\bearing$, corresponding to look-ahead vector $\lookahead$.
While a powerful control law, very high wind ratios degrade the performance and the ground speed based formulation does not handle \emph{excess wind} conditions (i.e. $\windratio\geq 1$) \cite{Furieri2017_windyNPFG}.
Noting any normal acceleration command in reality is applied about the aircraft's \emph{velocity-axis}, the reference acceleration may be reformulated about the airspeed vector
\begin{equation}
\label{eq:aNAref}
a_{A,\text{ref}}^N = k{v_A}^2\sin\errorangleA
\end{equation}
where $\errorangleA=\headingref-\heading \in\left[-\pi,\pi\right]$ is the angular error in heading $\heading$ from the heading reference $\headingref$, corresponding to air-mass relative look-ahead vector $\lookaheadA$.
In the following sections, the construction of the air-mass relative look-ahead vector $\lookaheadA$ is built up step-by-step: first considering a baseline, purely track-error based, nonlinear path following law, followed by feed-forward rotations both converting the ground relative bearing $\bearing$ to an air-mass relative heading reference $\xi_\text{ref}$ and considering the path curvature, and finally partitioning the control law into \emph{feasible} and \emph{infeasible} bearing cases while maintaining reference command continuity.
The approach follows closely to that in \cite{Furieri2017_windyNPFG}, with extensions/enhancements over the original formulation highlighted at each stage.
All vectors and rotations from the following development are displayed in Fig.~\ref{fig:geom}.
\subsection{Baseline path following}
\label{sec:path_following}
We consider a mathematical definition of the \emph{ideal tracking objective} described in Sec.~\ref{sec:bearing_feasibility}, i.e. the desired convergence behavior of the guidance law, assuming the bearing is feasible.
\begin{equation}
\text{ideal obj. }\begin{cases}
\underset{t\rightarrow\infty}{\lim}\mathbf{e}(t)=0 \\
\underset{t\rightarrow\infty}{\lim}\left(\hat{\mathbf{t}}_P(t)-\hat{\mathbf{v}}_G(t)\right)=0 \\
\underset{t\rightarrow\infty}{\lim}\left(\displaystyle \frac{d}{dt}\hat{\mathbf{t}}_P(t)-\frac{d}{dt}\hat{\mathbf{v}_G}(t)\right)=0
\end{cases}\label{eq:idealobj}
\end{equation}
where track-error $\mathbf{e}=\mathbf{p}-\mathbf{r}$, $\mathbf{r}$ is the vehicle position, $\hat{\mathbf{t}}_P$ is the path's unit tangent vector at the closest point $\mathbf{p}$, and $\hat{\mathbf{v}}_G=\mathbf{v}_G/\|\mathbf{v}_G\|, \|\mathbf{v}_G\|\neq 0$ is the unit ground speed vector.
In words, the vehicle should converge to the path while heading in the correct direction.
Similar to \cite{Cho2016_NPFG}, we define a purely track-error based look-ahead vector $\lookahead$:
\begin{equation}
\label{eq:look_ahead_vector}
\begin{array}{l}
\lookahead=\cos\theta_l\hat{\mathbf{e}} + \sin\theta_l\hat{\mathbf{t}}_P \\
\end{array}
\end{equation}
where $\hat{\mathbf{e}}=\mathbf{e}/\|\mathbf{e}\|, \|\mathbf{e}\|\neq0$ is the unit track-error and look-ahead angle $\theta_l=f\left(\mathbf{e}\right) \in\left[0,\frac{\pi}{2}\right]$ maps track-offset to a reference angle of approach to the path.
Shaping of the bearing transition through $\theta_l$ is achieved by combining the nominal mapping function found in \cite{Cho2016_NPFG} with a quadratic \emph{lead-in} curve, smoothly bringing in the desired change in $\bearing$ as the vehicle approaches the path:
\begin{equation}
\theta_l = \frac{\pi}{2}\left(1-\enorm\right)^2
\end{equation}
where $\enorm = \sat\left(\frac{\|\mathbf{e}\|}{\eb},0,1\right)$ is the normalized track-error within the track-error boundary $\eb$.
Similar to the guidance augmentation in~\cite{stastny2018_icuas_nmpc}, we extend the formulations in \cite{Furieri2017_windyNPFG,Cho2016_NPFG} with an adaptive track error boundary $e_b$, taking into account the current ground speed $v_G=\|\mathbf{v}_G\|$:
\begin{equation}
\eb = \begin{cases}
\frac{T_b}{2 \vgco} {v_G}^2 + \frac{T_b}{2} \vgco & v_G < \vgco \\
T_b v_G & \text{else}
\end{cases}
\end{equation}
where $T_b$ is a tunable look-ahead time constant, useful for modulating the time at which the aircraft begins turning into the path with respect to the approaching ground speed.
A ground speed cut-off $\vgco$ is incorporated within a piecewise quadratic function to smoothly saturate $e_b$ as $v_G\rightarrow 0$, avoiding singularities.
It is from a distance of $e_b$ the look-ahead vector will begin to transition from normal to tangent bearings, with respect to the path.
At this point, we need to augment $\lookahead$ for both wind and path curvature.
Look-ahead $\lookaheadA$ requires separate control laws for \emph{feasible} and \emph{infeasible} bearings, corresponding to the ideal and safety objectives, respectively.
\subsection{Feasible Bearing}
\label{sec:feasible_bearing}
The look-ahead vector $\lookahead$ describes a bearing necessary to drive convergence to a path with no curvature.
In the case that the bearing is \emph{feasible}, wind vector information may be utilized to translate appropriate heading references $\xi_\text{ref}$ necessary to achieve the ground relative motion defined by the bearing.
Towards this end, we rotate the ground-based look-ahead vector $\lookahead$ by angle $x$:
\begin{equation}
\label{eq:x}
x = \sin^{-1}\left(\windratio\sin\lambda\right)
\end{equation}
the resulting (curvature independent, i.e. $\curv=0$) heading vector reference then
\begin{equation}
\begin{array}{l}
\lookaheadAz = \mathbf{H}\left(x\right)\lookahead \label{eq:lAz} \\
\text{where}\quad\mathbf{H}\left(\cdot\right)=\left(\begin{matrix}
\cos\left\lbrace\cdot\right\rbrace & -\sin\left\lbrace\cdot\right\rbrace\\
\sin\left\lbrace\cdot\right\rbrace & \cos\left\lbrace\cdot\right\rbrace\end{matrix}\right)
\end{array}
\end{equation}
%

%
%
To further account for path curvature, we consider the ``on-track" wind triangle, i.e. angles $\lambda_0$, $x_0$, and $y_0$ and ground velocity $\mathbf{v}_{G_0}=v_{G_0}\hat{\mathbf{t}}_P$ at point $\mathbf{p}$, where
\begin{align}
\lambda_0 &= \operatorname{atan2}\left(\mathbf{w}\times\hat{\mathbf{t}}_P,\mathbf{w}\cdot\hat{\mathbf{t}}_P\right) \label{eq:lambda0} \\
x_0 &= \sin^{-1}\left(\windratio\sin\lambda_0\right) \label{eq:x0} \\
y_0 &= \pi - |x_0| - |\lambda_0| \label{eq:y0} \\
\vgz &= \sqrt{ {v_A}^2 + w^2 - 2 v_A w \cos y_0} \label{eq:vg0}
\end{align}
In this condition, imagining the vehicle is already tracking the path with $\|\mathbf{e}\|=0$ and $\mathbf{v}_A$ aligned with $\lookaheadAz$ (if calculated as in \eqref{eq:lAz} from the respective $x_0$), an additional normal acceleration
\begin{equation}
a_{G_0}^N={\vgz}^2\curv \label{eq:aNG0}
\end{equation}
is required to follow the path's curvature.
With a quasi-steady assumption on wind and differentiating \eqref{eq:y0},
\begin{align}
\dot{y}_0 &= -\dot{\lambda}_0-\frac{\windratio \cos\lambda_0 \dot{\lambda}_0}{\sqrt{1 + \left(\windratio\sin\lambda_0\right)^2}} \label{eq:ydot0_diff}
\end{align}
Noting the relationship between normal acceleration, linear speed, and angular speed $a^N=v\omega$, and following the guidance law in \eqref{eq:aNAref} it also holds:
\begin{align}
\dot{\lambda}_0 &=\frac{a_{G_0}^N}{\vgz} \label{eq:lambdadot0} \\
\dot{y}_0 &= k{v_A}\sin\eta_{c_0} \label{eq:ydot0}
\end{align}
where $\eta_{c_0}$ is the necessary additional rotation through which $\lookaheadAz$ must be transformed to obtain $\lookaheadA$.
Hence, plugging \eqref{eq:aNG0} into \eqref{eq:lambdadot0}, \eqref{eq:lambdadot0} into \eqref{eq:ydot0_diff}, then equating \eqref{eq:ydot0} and \eqref{eq:ydot0_diff}, we may obtain on-track curvature rotation:
\begin{equation}
\label{eq:etac0}
\begin{array}{l}
\eta_{c_0} = \\
\quad\sin^{-1}\left(\feas\left(\lambda_0,\windratio\right) \frac{\vgz\curv}{v_A k} \left(1 + \frac{\windratio\cos\lambda_0}{\sqrt{1 - \left(\windratio\sin\lambda_0\right)^2}}\right) \right)
\end{array}
\end{equation}
We highlight three enhancements present in the curvature rotation defined in \eqref{eq:etac0}, compared to that in \cite{Furieri2017_windyNPFG}:
\begin{itemize}
\item Considering only the on-track wind triangle avoids the necessity to saturate input to the arcsine function, previously required due to the mismatch between on-track angles and the aircraft centric wind triangle. 
\item We embed the on-track bearing feasibility $\feas\left(\lambda_0,\windratio\right)$ in order to zero-out the arcsine input argument as we approach the feasibility barrier, and thus that of an infeasible state where considering curvature no longer makes sense.
\item To avoid observed occasionally unintuitive flight trajectories caused by considering the path curvature when far from the track, an additional smooth limiter $\sigma_l=\sin^2\theta_l$ is included to bring in curvature adjustments only as we converge to the path: 
\end{itemize} 
\begin{align}
\eta_c &= \feas\left(\lambda,\windratio\right)\sigma_l\eta_{c_0} \label{eq:etac}
\end{align}
where $\feas\left(\lambda,\windratio\right)$ again zeros-out the consideration of path curvature as we approach the feasibility barrier; in this case in the aircraft centric frame to maintain reference continuity. 
With the final rotation necessary for tracking curvature, the airspeed relative look-ahead vector for the \emph{feasible} case may then be summarized as:
\begin{align}
\lookaheadAfeas &= \mathbf{H}\left(\eta_c\right)\lookaheadAz \\
\implies \lookaheadAfeas &= \mathbf{H}\left(x+\eta_c\right)\lookahead \label{eq:lAfeas}
\end{align}
noting here the rotation is one-dimensional, so the angles may be simply added.
\subsection{Infeasible Bearing}
\label{sec:infeasible_bearing}
\begin{figure}
\centering
\includegraphics[width=0.8\linewidth]{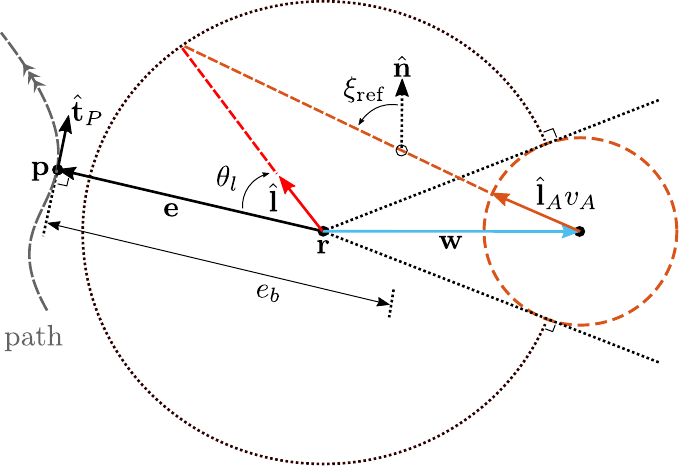}
\caption{Directional guidance -- geometry for infeasible bearing.}
\label{fig:geom_infeas}
\end{figure}

Following \cite{Furieri2017_windyNPFG}, when the bearing becomes fully infeasible, the look-ahead reference may be defined as (see also Fig.~\ref{fig:geom_infeas} for a visual geometric description):
\begin{equation}
\lookaheadAinfeas = \frac{\sqrt{w^2-{v_A}^2} \lookahead - \mathbf{w}}{\|\sqrt{w^2-{v_A}^2} \lookahead - \mathbf{w}\|} \label{eq:lAinfeas}
\end{equation}
This strategy considers the trade-off between tracking performance (ideal objective) and safety performance (safety objective) while $\lambda\in\left[\sin^{-1}\windratio^{-1},\pi\right]\cap\windratio\geq 1$, i.e. favoring a ``worst-case" safety configuration of facing against the wind as $\lambda\rightarrow\pi$, and that of the ideal objective defined in \eqref{eq:idealobj} when $\lambda$ resides at the feasibility barrier.
\begin{equation}
\text{safety obj. }\begin{cases}
\underset{t\rightarrow\infty}{\lim}a_{A,\text{ref}}^N(t)=0 \\
\underset{t\rightarrow\infty}{\lim}\hat{\mathbf{v}}_A(t)=-\hat{\mathbf{w}}(t) \\
\underset{t\rightarrow\infty}{\lim}\hat{\mathbf{e}}(t)=-\hat{\mathbf{w}}(t)
\end{cases}\label{eq:safetyobj}
\end{equation}
The latter of the requirements in \eqref{eq:safetyobj} correspond to the desire to minimize ``run-away" from the track.
Convergence analysis of the safety objectives \eqref{eq:safetyobj} defined for look-ahead law \eqref{eq:lAinfeas} may be found in \cite{Furieri2017_windyNPFG} which is similarly applicable to the present formulation.
Note, in this work, the infeasible look-ahead reference $\lookaheadAinfeas$ always uses the ``faster" heading solution of the two seen for the excess wind case in Fig.~\ref{fig:feas_cone}.

\subsection{Tuning}
The $k$ bounds for guaranteeing curvature convergence can be derived by considering the steady-state conditions $\mathbf{e}=\mathbf{0}$ in the ``worst-case" scenario $\hat{\mathbf{t}}_P=\hat{\mathbf{w}}$ (i.e. maximum required normal acceleration to maintain curvature), where $v_{G_0}=v_A+w$ and $\lambda_0=0$ are substituted within \eqref{eq:aNG0}, \eqref{eq:lambdadot0}, and \eqref{eq:ydot0_diff}:
\begin{equation}
\dot{y}_\text{ss} = -\left(v_A+w\right)|\curv|\left(\windratio+1\right) \label{eq:ydotss}
\end{equation}
Further considering the input argument of the arcsine function in \eqref{eq:etac0}, it may noticed that $k>\dot{y}_\text{ss}/v_A$ to ensure the equation is well defined, this resulting in the following $k$ bounds:
\begin{equation}
\begin{array}{l}
k>\left(1+\windratio\right)^2|\curv|
\end{array}\label{eq:k_bound}
\end{equation}
While the above initial analysis was also present in~\cite{Furieri2017_windyNPFG}, we handle a previously unconsidered practical implementation of this bound, that of potentially variable wind ratios $\windratio$ and path curvature $\curv$, in an adaptive way:
\begin{align}
k_\text{adj} &= k_\text{max} + \sigma_l\left(k - k_\text{max}\right) \label{eq:kadj} \\
k_\text{max} &= \begin{cases}
\operatorname{max}\left(k, k_\text{mult}\left(1+\windratio\right)^2|\curv|\right) & \windratio\geq 1 \\
\operatorname{max}\left(k, 4 k_\text{mult}|\curv|\right) & \text{else}
\end{cases}
\end{align}
where $k_\text{max}$ is the maximum of the operator defined proportional gain $k$ and the minimum required gain from \eqref{eq:k_bound} (with some tolerance, $k_\text{mult}$), and $k_\text{adj}$ is the resulting adjusted gain used by the controller.
Note $\left(1+\left(\windratio=1\right)\right)^2 = 4$, which is held as a constant multiplier in the $\windratio<1$ case. 
This logic alleviates the need for condition specific tuning and ensures convergence is maintained while still allowing operator defined dynamics of the control response whenever the bounds are not exceeded.
\subsection{Control allocation}
The commanded lateral acceleration from the directional guidance (following the control law in~\eqref{eq:aNAref}) is translated into a roll angle reference via the common coordinated turn assumption: $\phi_\text{ref} = \tan^{-1}\left(a_{A,\text{ref}}^N/g\right)$, where $g$ is the acceleration of gravity.
It is then the task of the lower-level control loops to track this reference, see Sec.~\ref{sec:flight_experiments} for control architecture details.
%

\section{AIRSPEED REFERENCE COMPENSATION}
\label{sec:airspeed_reference_compensation}
In this section, we extend the high-level guidance logic by adding an additional control, the airspeed reference, developing an energy efficient \emph{airspeed compensation} logic.
With the assumption that extra commanded airspeed entails extra energy usage, we wish to only increase the reference as much as necessary to prevent run-away (stay on track) until winds have receded.
I.e. when the feasibility barrier has been crossed, an ideal equilibrium point of $v_G=\|\mathbf{e}\|=0$ should be approached.
The following subsections develop successive stages of compensation logic; namely, \emph{wind excess regulation}, \emph{track keeping}, and \emph{minimum ground speed maintenance}.
\subsection{Wind Excess Regulation}
To achieve the first component of the desired equilibrium, $v_G=0$, we define a positive speed increment $\Deltavaw$ corresponding to the difference between the wind speed $w$ and nominal reference airspeed $\vanom$, i.e. the \emph{excess wind speed} $\Delta w = \sat \left(w - \vanom, 0, \Deltavamax\right)$, where $\Deltavamax = \operatorname{max} \left(\vamax - \vanom, 0\right)$ is the maximum allowed airspeed reference increment, derived from the maximum available airspeed setting $\vamax$:
\begin{equation}
\Deltavaw = \Delta w \left(1 - \feas\left(\lambda,\windratio\right)\right) \label{eq:deltavaw}
\end{equation}
This wind excess -based speed increment may be added to the nominal reference $\vanom$ towards regulating $v_G\rightarrow 0$, though small perturbations will induce small steady-state tracking errors which may only grow over time, unless wind speeds recede.
\subsection{Track Keeping}
\label{sec:track_keeping}
To further stay on track in excess wind speeds, an additional speed increment $\Delta v_A^e$ corresponding to the normalized track-error $\enorm$ may be defined:
\begin{equation}
\Deltavae = \Deltavaemax k_{\enorm} k_w \left(1 - \feas\left(\lambda,\windratio\right)\right)
\end{equation}
where $\Deltavaemax$ is the maximum allowed speed increment generated from track-error.
The gains $k_{\enorm}$ and $k_w$ are used to tune track-error and wind speed excess derived saturation ramps:
\begin{align}
k_{\enorm} &= \sat\left( \frac{\enorm}{\Deltaebuf}, 0, 1\right) \\
k_w &= \sat\left( \frac{\Delta w}{\Deltawbuf}, 0, 1\right)
\end{align}
$k_{\enorm}$ is scaled by a chosen fraction of the normalized track-error $\Deltaebuf$, setting the proximity at which $\Deltavaemax$ is applied in full, while $k_w$ is scaled by $\Deltawbuf$ to ensure no airspeed increment is applied in the condition that the feasibility function lies within the extended buffer zone below $\windratio=1$.
The track offset -based speed increment $\Deltavae$ assists $\Deltavaw$ by increasing the airspeed enough to overpower the current wind speed, returning the aircraft to the path, at which point the term again zeros out.
With both increments in play, the augmented airspeed reference combines them as follows:
\begin{equation}
\label{eq:va_ref}
\varef = \vanom + \operatorname{min}\left( \Deltavaw + \Deltavae, \Deltavamax\right)
\end{equation}
\subsection{Maintaining a Minimum Forward Ground Speed}
\label{sec:min_fwd_gsp}
\begin{figure}
\centering
\includegraphics[width=\linewidth]{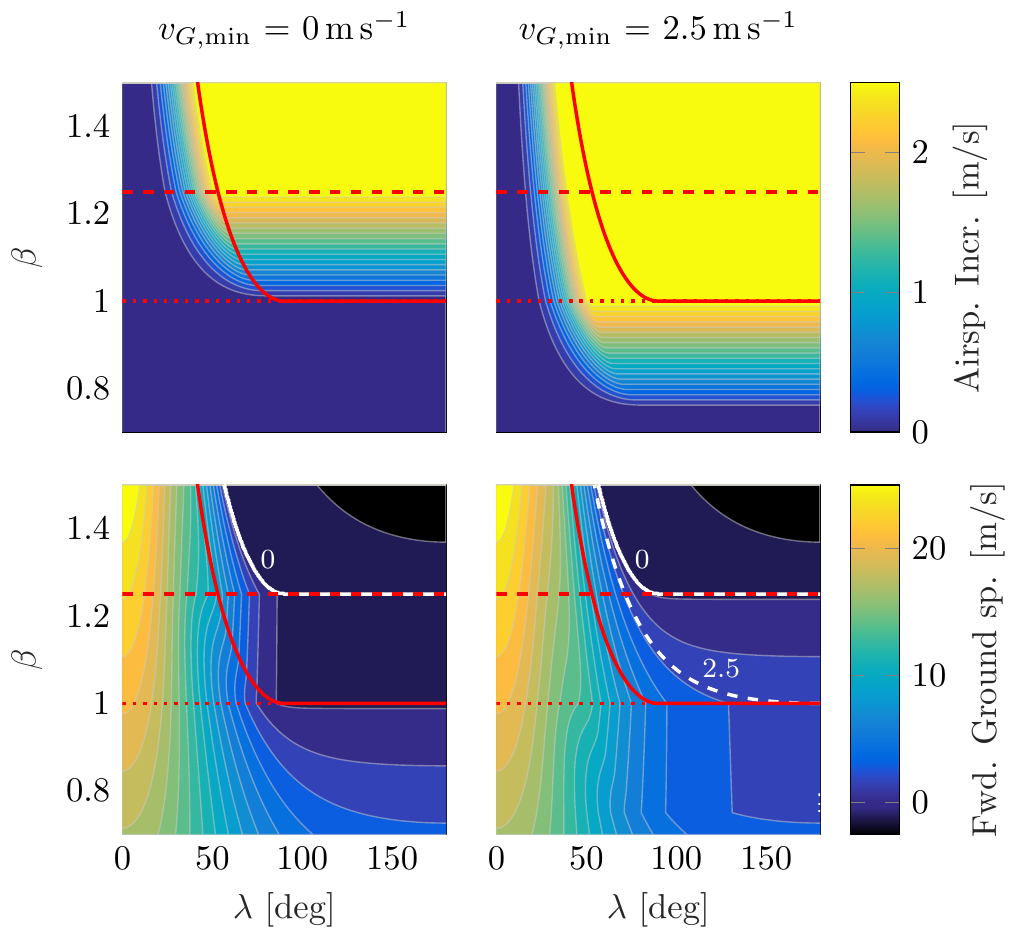}
\caption{Airspeed reference compensation and resulting forward ground speed with (right) and without (left) a minimum ground speed applied (note no track-offset increment is added here). For illustrative purposes, the look-ahead logic defined in Sec.~\ref{sec:dir_guidance} is assumed perfectly tracked (or in steady-state condition). $v_{A,\text{nom}}=$\SI{10}{\meter\per\second} (dotted red line) and $v_{A,\text{max}}=$\SI{12.5}{\meter\per\second} (dashed red line). The solid red line indicates the feasibility boundary. Note airspeed is only incremented as necessary to achieve the desired forward ground speed until reaching the upper saturation bound of $\Delta v_{A,\text{max}}$.}
\label{fig:airsp_incr}
\end{figure}

Though the incremented airspeed reference in \eqref{eq:va_ref} will maintain zero track-error, it may further be desirable that forward progress is made on a given path -- e.g. if the vehicle should attempt to return home, or at the least complete its current set of mission waypoints.
Towards this end, the bearing feasibility function can be utilized for efficient tracking of an operator-set minimum  \emph{forward ground speed} $\vgmin$, with forward ground speed defined as the 2D (horizontal) projection onto the airspeed vector $v_{G,\text{fwd}}=\mathbf{v}_G\cdot\frac{\mathbf{v}_A}{\|\mathbf{v}_A\|}$. 
$\vgmin$ may then be used to augment the wind ratio and, further, airspeed increments, effectively imitating a higher wind speed which the logic must compensate: 
\begin{align}
\windratio_G &= \frac{w + \vgmin}{v_A} \label{eq:windratioG} \\
\Delta w &= \sat \left(w - \vanom + \vgmin, 0, \Deltavamax\right) \label{eq:deltawG} \\
\Deltavaw &= \Delta w \left(1 - \feas\left(\lambda,\windratio_G\right)\right) \label{eq:deltavawG}
\end{align}
Note that with a minimum ground speed defined, the track keeping logic defined in Sec.~\ref{sec:track_keeping} should be disabled, i.e. $\Deltavae=0$.
Fig.~\ref{fig:airsp_incr} illustrates the feed-forward airspeed reference mapping resulting from the minimum ground speed logic.
%

\section{FLIGHT EXPERIMENTS}
\label{sec:flight_experiments}
In this section, we present flight demonstrations on a small (\SI{1.8}{\meter} wingspan, \SI{1.3}{\kilo\gram}), low-speed test platform, Easyglider (see Fig.~\ref{fig:teaser}), from two mountainous regions in Switzerland, showcasing the performance of the developed guidance laws in strong winds.
The guidance algorithm has been programmed in C/C++ on a \textit{Pixhawk} autopilot (\SI{168}{\mega\hertz} Cortex-M4F microcontroller with \SI{192}{\kilo\byte} RAM) running \emph{PX4}\footnote{\url{http://dev.px4.io}} firmware.
PX4 implementations of a cascaded PID-based attitude/rate control (with feed-forward turn compensation), airspeed and altitude control via Total Energy Control System (TECS)~\cite{tecs1987}, and an online Extended Kalman Filter (EKF) are utilized for tracking guidance commands and feeding back state estimates, respectively.
We note that all underlying control and estimation structures are operational with a standard low-cost sensor suite for small fixed-wing UAVs (see exemplary sensor selection in~\cite{oettershagen_jfr2018}) and further require no model-based assumptions.
Guidance parameters held constant for both flights may be found in Table~\ref{tab:guidance_parameters}.
All displayed airspeeds are ``true" airspeeds (TAS), i.e. relative to the airmass. 
To keep wind, ground speed, and TAS inputs to the guidance algorithm compatible with eachother, a ``filtered" TAS estimate is obtained by subtracting the wind estimate from the GNSS velocity.
As the wind estimate is already filtered, this further smooths out the typically noisy airspeed measurements (from a pitot-static tube with differential and ambient pressure sensors) that would otherwise degrade the guidance commands.  
\begin{table}[h]
\centering
	\caption{Guidance parameters used in flight experiments.}
	\begin{tabular}{!{\vrule width 0.5pt}ccc!{\vrule width 0.5pt}ccc!{\vrule width 0.5pt}}
	\noalign{\hrule height 0.5pt}
	Param & Value & Unit & Param & Value & Unit \\\noalign{\hrule height 0.5pt}
	$\beta_\text{buf}$ & 0.1 & - - & $\vanom$ & 8.8 & \SI{}{\meter\per\second}\\
	$\lambda_\text{co}$ & 1.0 & \SI{}{\degree} & $\vamax$ & 15.0 & \SI{}{\meter\per\second}\\
	$\vgco$ & 1.0 & - - & $\Deltaebuf$ & 0.5 & - - \\
	$T_b$ & 7.0 & \SI{}{\second} & $\Deltawbuf$ & 0.5 & \SI{}{\meter\per\second}\\
	$k$ & 0.11 & - - & $\Deltavaemax$ & 3.0 & \SI{}{\meter\per\second}\\
	$k_\text{mult}$ & 1.1 & - - &  &  & \\\noalign{\hrule height 0.5pt}
	\end{tabular}
	\label{tab:guidance_parameters}
\end{table}
\subsection{Wind Excess Regulation and Minimum Ground Speed Maintenance}
\begin{figure}
  \includegraphics[width=\linewidth]{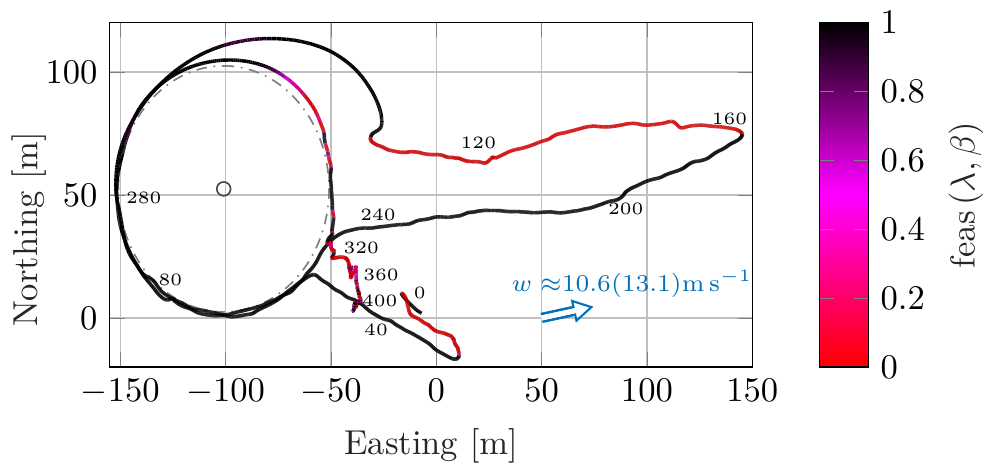}
  \caption{Flight experiment: Lamboing, Switzerland (892 AMSL - Plateau de Diesse). Aircraft position colored with the current bearing feasibility. Mean wind speed was \SI{10.6}{\meter\per\second}, gusting to \SI{13.1}{\meter\per\second} during the flight period.}
  \label{fig:flight_exp-lamboing_position}
\end{figure}

\begin{figure*}[tb]
  \includegraphics[width=0.95\textwidth]{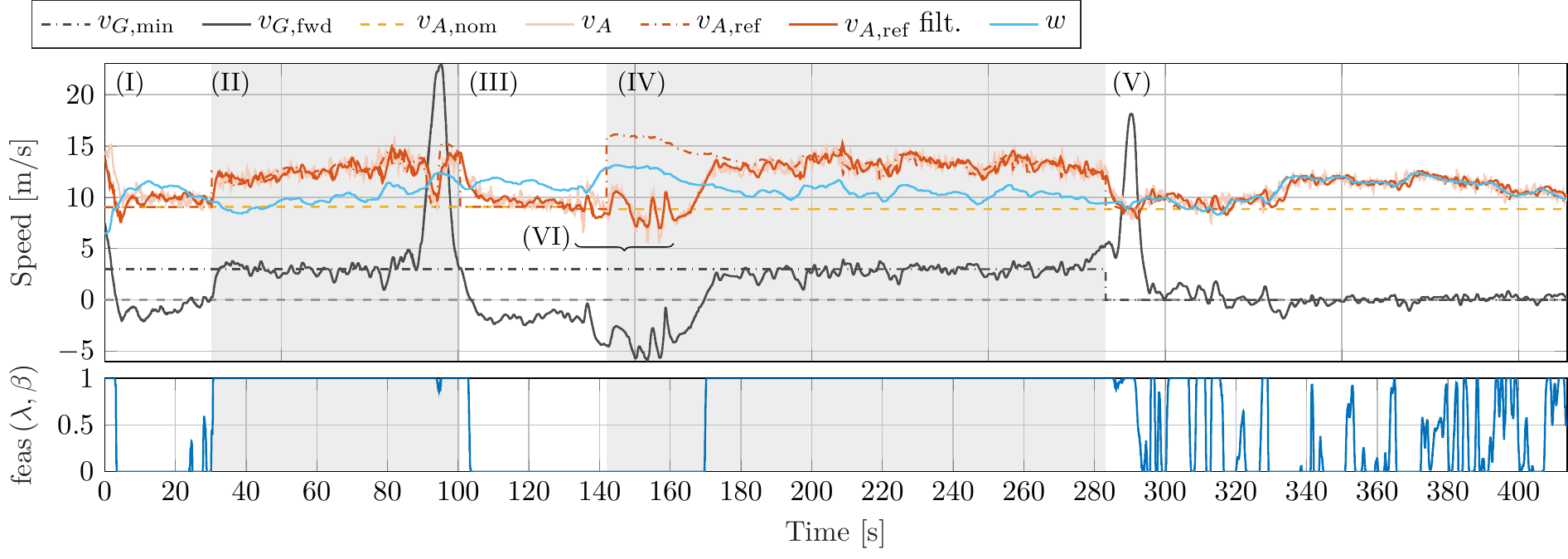}\par
  \caption{Flight experiment: Lamboing, Switzerland (892 AMSL - Plateau de Diesse). Speeds and bearing feasibility. \Romannum{1}. Automatic (backwards) take-off, airspeed compensation disabled, \Romannum{2}. Approach and tracking of loiter circle with $\vgmin=$\SI{3}{\meter\per\second}, \Romannum{3}. Airspeed compensation disabled, leading to run-away bottoming out near $v_{G,\text{fwd}}=$\SI{-5}{\meter\per\second}, \Romannum{4}. $\vgmin=$\SI{3}{\meter\per\second} re-enabled, \Romannum{5}. Wind excess regulation is demonstrated by setting $\vgmin=$\SI{0}{\meter\per\second}, \Romannum{6}. Stall due to turbulence and gust increase while flying with negative ground speed.}
  \label{fig:flight_exp-lamboing_states}
\end{figure*}

Figures~\ref{fig:flight_exp-lamboing_position} and~\ref{fig:flight_exp-lamboing_states} show the position and state trajectories of a fully automatic ca. \SI{7}{\minute} flight experiment on a windy plateau in the Jura Mountain range in Switzerland.
Stages (\Romannum{1})-(\Romannum{5}) in Fig.~\ref{fig:flight_exp-lamboing_states} step through various compensated and uncompensated guidance modes, however no track keeping is enabled.
A clear reduction in run-away can be seen in Fig.~\ref{fig:flight_exp-lamboing_position} when comparing the uncompensated case (\Romannum{3}: $t=$101-142\SI{}{\second}) to that of wind excess regulation (\Romannum{5}: $t=$283-414\SI{}{\second}).
Further, when commanded, minimum forward ground speed is maintained with a mean undershoot of \SI{0.51}{\meter\per\second} (with one standard deviation error \SI{1.07}{\meter\per\second}), increasing the airspeed reference when the bearing would otherwise become infeasible, and decreasing appropriately on the down wind legs, see (\Romannum{2}), (\Romannum{4}), and (\Romannum{5}); this, however, with the exception of the stall denoted by (\Romannum{6}).
At $t=$\SI{135}{\second}, while the Easyglider is running away at up to \SI{-5}{\meter\per\second} forward ground speed (due to the disabled airspeed compensation logic), the aircraft encounters some turbulence (note the abrupt large airspeed fluctuations) followed by a gust increase of \SI{2.7}{\meter\per\second}.
These effects induce a stall from which the lower-level control loops spend the next several seconds regaining control.
Once lower-level stabilization is regained, tracking of the guidance commands quickly resumes.
This momentary lapse in low-level stabilization highlights the risk of flying small UAVs in such conditions and further motivates consideration of coupled longitudinal effects (e.g. angle of attack) within future iterations of wind-robust guidance and control.
\subsection{Track Keeping}
\begin{figure}
  \includegraphics[width=\linewidth]{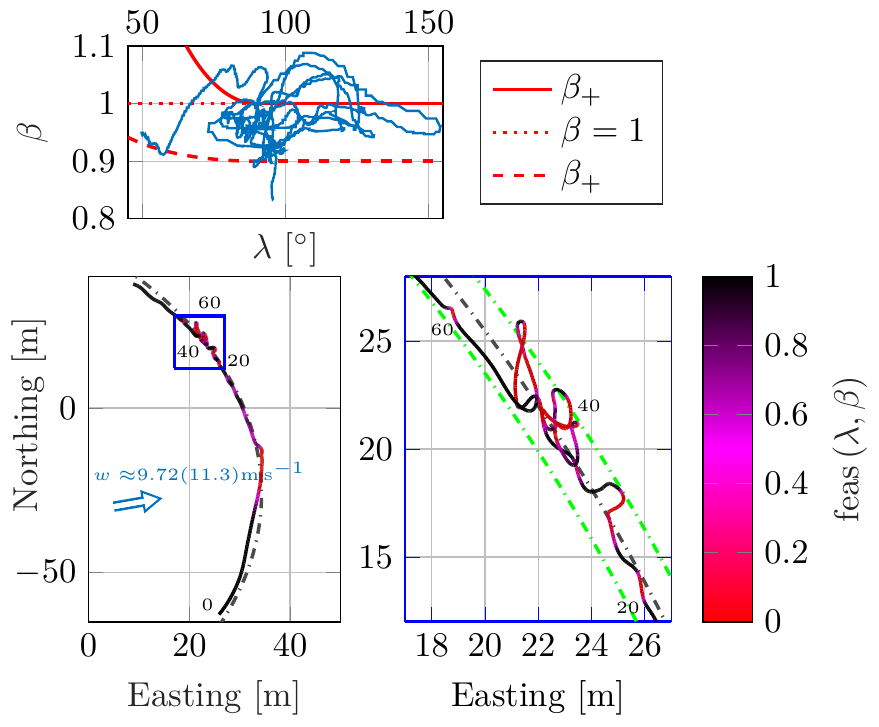}
  \caption{Flight experiment: Uetliberg, Switzerland (943 AMSL - mountain top). Aircraft position colored with the current bearing feasibility. Green dashed-dot lines within the zoom-in plot (right) indicate \SI{1}{\meter} bounds on the path. Mean wind speed was \SI{9.72}{\meter\per\second}, gusting to \SI{11.3}{\meter\per\second} during the flight period. Wind ratio vs. the angle between wind and bearing vectors over feasibility function boundaries (top).}
  \label{fig:flight_exp-uetliberg_position}
\end{figure}

\begin{figure}
  \includegraphics[width=\linewidth]{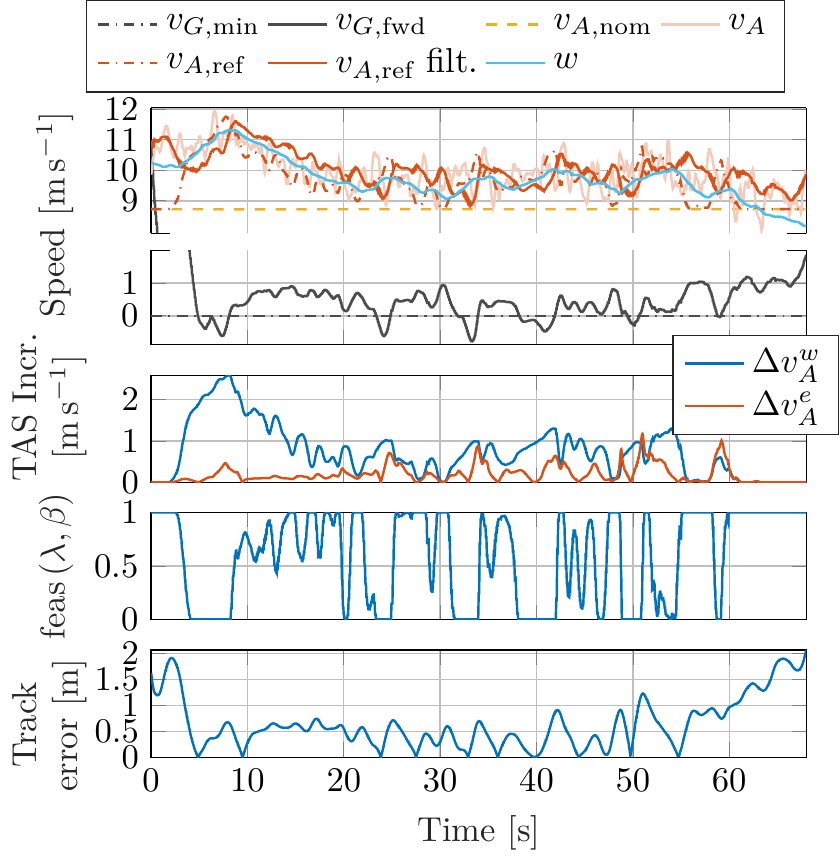}
  \caption{Flight experiment: Uetliberg, Switzerland (943 AMSL - mountain top). Aircraft states for a track keeping experiment in strong ridge updrafts and horizontal winds.}
  \label{fig:flight_exp-uetliberg_states}
\end{figure}

Figures~\ref{fig:flight_exp-uetliberg_position} and~\ref{fig:flight_exp-uetliberg_states} show position and state trajectories for a portion of a flight experiment conducted on the ridge of Uetliberg Mountain, Switzerland where the track keeping mode is enabled.
Figure~\ref{fig:flight_exp-uetliberg_position} details a full \SI{40}{\second} in which the guidance holds the aircraft at near zero ground speed with less than \SI{1}{\meter} track error, despite facing into gusting winds nearly constantly above the nominal airspeed.
In Fig.~\ref{fig:flight_exp-uetliberg_position}, $\Deltavaw$ can be seen effectively tracking the wind excess, while $\Deltavae$ adjusts for deviations from the track caused by smaller turbulences and/or gusts.
The importance of the bearing feasibility function's buffer zone is further highlighted here, as $\beta$ stays near 1, and $\lambda$ near \SI{90}{\degree}.
\section{DISCUSSION \& FUTURE WORK}
In this work, we developed and demonstrated a novel guidance law capable of preventing run-away of small, low-speed fixed-wing UAVs flying in strong winds.
Though the controller was shown effective in flight experiments, one particularly noteworthy observation during testing was that care should be taken in tuning of noise values for the wind estimates within the EKF.
Though highly dependent on the given estimation formulation, a general guidance may be taken that too slow or fast tuning of wind estimation responses causes, respectively, too slow or fast reactions the airspeed compensation logic, leading in the prior case to steady state tracking errors, and in the latter case to noisy, oscillatory guidance commands. A trade-off between the scale of gusts one wishes to capture vs the performance of the controller should be weighed.
It is further apparent that flight within very turbulent conditions will require future work on coupled wind-robust algorithms considering longitudinal lower-level dynamics of the UAV.
The authors lastly note that, though not shown here for brevity, the developed guidance law has further been applied/tested for straight line and ellipse following.

%




\section*{ACKNOWLEDGMENT}
This research received funding from the Federal office armasuisse S+T under project number n\SI{}{\degree}050-45 and the ETH Foundation Grant ETH-12 16-2 (Sun2Ice Project).
%


\bibliographystyle{ieeetr}
\bibliography{lib.bib}

\begin{thebibliography}{10}

\bibitem{oettershagen_jfr2018}
P.~Oettershagen, T.~Stastny, T.~Hinzmann, K.~Rudin, T.~Mantel, A.~Melzer,
  B.~Wawrzacz, G.~Hitz, and R.~Siegwart, ``Robotic technologies for
  solar-powered uavs: Fully autonomous updraft-aware aerial sensing for
  multiday search-and-rescue missions,'' {\em Journal of Field Robotics},
  vol.~35, no.~4, pp.~612--640, 2018.

\bibitem{warren2015_emergencylanding}
M.~Warren, L.~Mejias, J.~Kok, X.~Yang, F.~Gonzalez, and B.~Upcroft, ``An
  automated emergency landing system for fixed-wing aircraft: Planning and
  control,'' {\em Journal of Field Robotics}, vol.~32, no.~8, pp.~1114--1140,
  2015.

\bibitem{Klein2017_emergencylanding_dubinswind}
M.~{Klein}, A.~{Klos}, J.~{Lenhardt}, and W.~{Schiffmann}, ``Wind-aware
  emergency landing assistant based on dubins curves,'' in {\em 2017 Fifth
  International Symposium on Computing and Networking (CANDAR)}, pp.~546--550,
  Nov 2017.

\bibitem{Marina2017_vectorfieldpaparazi}
H.~G. {de Marina}, Y.~A. {Kapitanyuk}, M.~{Bronz}, G.~{Hattenberger}, and
  M.~{Cao}, ``Guidance algorithm for smooth trajectory tracking of a fixed wing
  uav flying in wind flows,'' in {\em 2017 IEEE International Conference on
  Robotics and Automation (ICRA)}, pp.~5740--5745, May 2017.

\bibitem{beard2014pathfollowwind}
R.~W. Beard, J.~Ferrin, and J.~Humpherys, ``Fixed wing uav path following in
  wind with input constraints,'' {\em IEEE Transactions on Control Systems
  Technology}, vol.~22, pp.~2103--2117, Nov 2014.

\bibitem{Park2007_stability}
{S. Park, J. Deyst and J. P. How}, ``Performance and lyapunov stability of a
  nonlinear path following guidance method,'' {\em Journal of Guidance,
  Control, and Dynamics}, vol.~30, no.~6, pp.~1718--1728, 2007.

\bibitem{Curry2013_L2plus}
R.~Curry, M.~Lizarraga, B.~Mairs, and G.~H. Elkaim, ``L2+, an improved line of
  sight guidance law for {UAV}s,'' in {\em 2013 American Control Conference},
  pp.~1--6, June 2013.

\bibitem{Furieri2017_windyNPFG}
L.~Furieri, T.~Stastny, L.~Marconi, R.~Siegwart, and I.~Gilitschenski, ``Gone
  with the wind: Nonlinear guidance for small fixed-wing aircraft in
  arbitrarily strong windfields,'' in {\em 2017 American Control Conference
  (ACC)}, pp.~4254--4261, May 2017.

\bibitem{Cho2016_NPFG}
N.~Cho, Y.~Kim, and S.~Park, ``Three-dimensional nonlinear differential
  geometric path-following guidance law,'' {\em Journal of Guidance, Control,
  and Dynamics}, 2015.

\bibitem{stastny2018_icuas_nmpc}
T.~{Stastny} and R.~{Siegwart}, ``Nonlinear model predictive guidance for
  fixed-wing uavs using identified control augmented dynamics,'' in {\em 2018
  International Conference on Unmanned Aircraft Systems (ICUAS)}, pp.~432--442,
  June 2018.

\bibitem{tecs1987}
K.~R. Bruce, J.~R. Kelly, and L.~H. Person, ``{NASA} {B737} flight test results
  of the total energy control system,'' {\em AIAA Guidance Navigation and
  Control (GNC) Conference}, 1987.

\end{thebibliography}

\end{document}